\documentclass[conference]{IEEEtran}
\IEEEoverridecommandlockouts
\usepackage{booktabs} 
\usepackage{times}
\usepackage{epsfig}
\usepackage{amsmath}
\usepackage{amssymb}
\usepackage{enumerate}
\usepackage{graphicx}
\usepackage{adjustbox}
\usepackage{indentfirst}
\usepackage{array}
\usepackage{multirow}
\usepackage{float}
\usepackage{hhline}
\usepackage{bm}
\usepackage{subfigure}
\usepackage{tabularx}
\usepackage{color}
\usepackage{url}
\usepackage{fancyhdr}
\def\BibTeX{{\rm B\kern-.05em{\sc i\kern-.025em b}\kern-.08em    T\kern-.1667em\lower.7ex\hbox{E}\kern-.125emX}}

\fancypagestyle{firstpage}{%
  \lhead{Accepted in 2019 International Joint Conference on Neural Networks (IJCNN)}
  \lfoot{\small{\copyright2019 IEEE. Personal use of this material is permitted. Permission from IEEE must be obtained for all other users, including reprinting/republishing this material for advertising or promotional purposes, creating new collective works for resale or redistribution to servers or lists, or reuse of any copyrighted components of this work in other works.}}
}

\begin{document}

\title{Spontaneous Facial Micro-Expression Recognition using 3D Spatiotemporal Convolutional Neural Networks}

\author{\IEEEauthorblockN{Sai Prasanna Teja Reddy\IEEEauthorrefmark{1},
Surya Teja Karri\IEEEauthorrefmark{2},
Shiv Ram Dubey\IEEEauthorrefmark{3} and 
Snehasis Mukherjee\IEEEauthorrefmark{4}}
\IEEEauthorblockA{Computer Vision Group,
Indian Institute of Information Technology, Sri City, India\\
\IEEEauthorrefmark{1}prasannateja.b15@iiits.in,
\IEEEauthorrefmark{2}suryateja.k15@iiits.in,
\IEEEauthorrefmark{3}srdubey@iiits.in}
\IEEEauthorrefmark{4}snehasis.mukherjee@iiits.in}

\maketitle
\thispagestyle{firstpage}

\begin{abstract}
Facial expression recognition in videos is an active area of research in computer vision. However, fake facial expressions are difficult to be recognized even by humans. On the other hand, facial micro-expressions generally represent the actual emotion of a person, as it is a spontaneous reaction expressed through human face. Despite of a few attempts made for recognizing micro-expressions, still the problem is far from being a solved problem, which is depicted by the poor rate of accuracy shown by the state-of-the-art methods. A few CNN based approaches are found in the literature to recognize micro-facial expressions from still images. Whereas, a spontaneous micro-expression video contains multiple frames that have to be processed together to encode both spatial and temporal information. This paper proposes two 3D-CNN methods: MicroExpSTCNN and MicroExpFuseNet, for spontaneous facial micro-expression recognition by exploiting the spatiotemporal information in CNN framework. The MicroExpSTCNN considers the full spatial information, whereas the MicroExpFuseNet is based on the 3D-CNN feature fusion of the eyes and mouth regions.
The experiments are performed over CAS(ME)$\widehat{}^2$ and SMIC micro-expression databases. The proposed MicroExpSTCNN model outperforms the state-of-the-art methods.
\end{abstract}

\section{Introduction}
Facial micro-expressions are subtle and involuntary. A facial micro-expression is a stifled facial expression that lasts only for a very brief interval of time (i.e., $40$ milliseconds) \cite{ekman2009lie}. They are the result of either conscious suppression or unconscious repression of expressions. Recognizing a micro-expression with the human eye is an extremely difficult task. Understanding the micro-expressions helps us to identify the deception and to understand the true mental condition of a person. The automatic micro-expression recognition through videos is very challenging as it is present in very few early frames of the video \cite{takalkar2018survey}. Initially, the expression recognition problem was mainly solved by the matching of hand-designed micro-expression descriptors extracted from the images/videos \cite{mukherjee2016recognizing}, \cite{li2018towards}. In recent years, the researchers have explored few deep learning based methods for micro-expression recognition \cite{merghani2018review}.

The traditional hand-crafted feature based methods for analyzing micro-expressions include spatiotemporal local binary pattern (LBP) \cite{huang2015facial}, LBP-TOP \cite{wang2014lbp}, directional mean optical flow feature \cite{liu2016main}, etc. However, the main downside with these methods is due to the extraction of mostly superficial information from the video and lack of required information for abstract feature representation.

Recently, deep learning based methods like convolutional neural networks (CNN) have gained popularity and widely being used to solve various computer vision problems \cite{goodfellow2016deep} including Image Classification \cite{krizhevsky2012imagenet}, Semantic Segmentation \cite{chen2018deeplab}, Blind Image Quality Assessment \cite{hou2015blind}, Face Anti-spoofing \cite{nagpal2018performance}, Routine Colon Cancer Nuclei Classification \cite{basha2018rccnet} and many more. In general, the deep learning based techniques are observed to outperform the hand-crafted techniques in most of the computer vision problems. Recently, a few CNN based approaches are proposed for micro-expression recognition \cite{guo2017multi,liong2018,cmu,patel2016selective}. However, the conventional deep learning techniques proposed for facial micro-expression recognition use CNN, RNN and/ or combinations of CNN and RNN. These approaches generally use CNN to extract the spatial feature for each frame and feed to RNN to encode the temporal correlation between the frames in the expression video. Thus, these methods are not able to encode the spatiotemporal relationship between the video features simultaneously. In order to overcome the limitations of the existing techniques, we propose two 3D CNN models (MicroExpSTCNN and MicroExpFuseNet) which extract both the spatial and temporal information simultaneously by applying the 3D convolution operation over the video.

The main contributions of this paper are summarized as:
\begin{enumerate}
    \item A MicroExpSTCNN model is proposed which extracts both spatial and temporal features of the expression video for classification. We have achieved the state-of-the-art performance over benchmark micro-expression datasets using the proposed MicroExpSTCNN model.
    \item A two stream MicroExpFuseNet model is proposed to combine the features extracted from the eyes and mouth regions only.
    \item Experiments are conducted on the intermediate and late fusion of eyes and mouth regions based 3D-CNNs.
    \item The effect of different facial features is analyzed using the salience maps.
    \item The effect of varying 3D kernel sizes has also been experimented for micro-expression recognition.
\end{enumerate}

This paper is structured as follows: Section II presents a detailed review of literature on micro-expression recognition. Section III proposes the proposed MicroExpSTCNN and MicroExpFuseNet architectures; Section IV summarizes the experimental settings followed by a detailed discussion on the experimental results and analysis in Section V. Finally, Section VI concludes the paper with remarks.

\section{Literature Review}
In this section, we briefly review and discuss the state-of-the-art approaches for micro-expression recognition. This literature review has two sub-sections: we first start our discussions on hand-designed methods followed by the recent learning based methods.
\subsection{Hand-Designed Methods}
Hand-designed feature based approaches for micro-expression recognition were started a decade back. Wu et al. \cite{wu2011machine} have designed an automatic system to locate the face. They used the Gabor filters to extract the features and Gentleboost as the feature selector. They used Support Vector Machine (SVM) over the generated features for recognizing the facial micro-expressions. Polikovsky et al. \cite{polikovsky2009facial} have divided the face into several sub-regions. Motion information in each sub-region is represented using the 3D-Gradient orientation histogram descriptor to capture the correlation between the frames. Lu et al. \cite{lu2018} have proposed a feature descriptor to uniquely identify the expression by fusing the motion boundary histograms. The feature descriptor is generated by fusing the differentials of horizontal and vertical components of optical flow. The generated feature vector is fed into SVM for classification of micro-expressions. Shreve et al. \cite{shreve2011macro} have proposed a unified method to recognize both the macro and micro facial expressions in a video using spatio-temporal strain on the face caused by the non-rigid motions. They have calculated the strain magnitude from different facial regions such as chin, mouth, cheek, forehead, etc. to distinguish between the macro and micro expressions. Pfister et al. \cite{pfister2011recognising} have used the spatio-temporal local texture descriptors and various classifiers to recognize the micro-expressions. They proposed a temporal interpolation model to intercept the problem of variable video lengths. Zhao et al. have proposed the Local Binary Pattern - Three Orthogonal Planes (LBP-TOP) feature for facial expression recognition using dynamic texture \cite{zhao2007dynamic} to exploit the spatio-temporal information in very compact form. Wang et al. \cite{wang2015dynamic} have proposed Local Binary Pattern with Six Intersection Points to tackle the problem of redundancy from the LBP-TOP. They fed the extracted features into SVM classifier for facial expression recognition. In practice, the LBP-TOP lacks with the sufficient features as video is represented by three orthogonal frames. Huang et al. \cite{huang2016spontaneous} have proposed the spatio-temporal completed local quantized patterns which compute the vector quantization and use the codebook to learn more dynamic patterns. Recently, Liu et al. \cite{liu2016main} have used the optical flow field computed over different sub-regions of the face image to recognize the facial micro-expression. In order to reduce the dimensionality of the feature vector, they have computed the  directional mean optical flow feature. The feature vector is then fed into the SVM classifier for training and recognition of micro-expressions and provided better results than other hand-crafted methods. 
\begin{figure*}[!t]
\centering
\includegraphics[width=\textwidth]{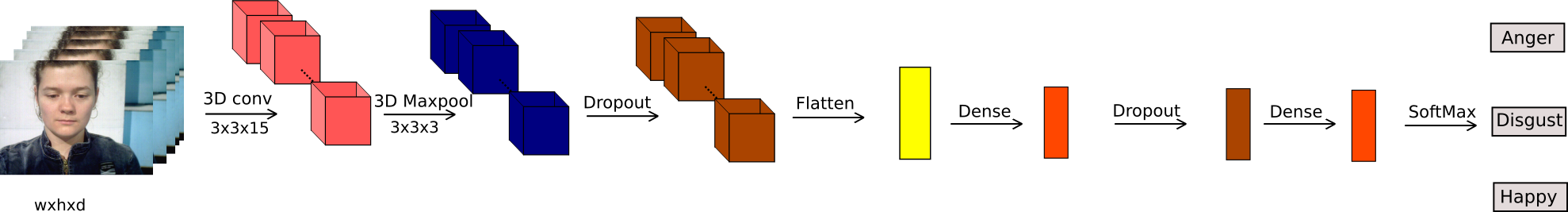}
\caption{The proposed MicroExpSTCNN architecture for micro-expression recognition using 3D-CNN.}
\label{fig:model1}
\end{figure*}

\subsection{Learning-Based Methods}
Recent advances in GPU based computing facilities have made possible to train the deep learning models with large datasets. The learning based methods have been applied to many vision problems, including classification, segmentation, detection, etc. Recently, efforts are being made by the computer vision scientists to apply deep learning models for micro-expression recognition. Grobava et al. \cite{grobova2017} have used the facial landmarks as the input vectors and applied the machine learning techniques like Support Vector Machine and Random Forest classifiers to classify the micro-expressions. Li et al. \cite{li2016spontaneous} have used the multi-task learning to identify the facial landmarks. The facial region is divided into facial sub-regions and region of interest based histogram of oriented optical flow is used to represent the micro-expression features. The Convolutional Neural Network (CNN) is the recent trend in deep learning to solve vision problems where images or videos are used as the input. Liong et al. \cite{liong2018} have proposed a method to classify the micro-expression by extracting the optical flow feature from the reference frame of a micro-expression video. Then, the extracted optical flow features are fed into a 2D CNN model for expression classification. Takalkar et al. \cite{takalkar2017} have utilized the data augmentation techniques to generate the synthetic images which are used to train the deep convolutional neural network (DCNN) to classify the micro-expressions. Li et al. \cite{li2018} have detected the facial sub-regions using a multi-task convolutional network. These facial sub-regions contain the important motion features required for classification. They used a fused deep convolutional network to estimate the optical flow features of the micro-expression. The obtained optical flow features are then fed into the SVM for training and recognition. Mayya et al. \cite{veena2016} have performed the temporal interpolation over the original video sequences. The output of interpolation is fed as an input to deep convolutional network for expression classification. The drawback of this method is that it may lose the important temporal information. Wang et al. \cite{wang2018} have used the transfer learning and residual network as a baseline architecture. They have further used the micro attention units that specifically learns the micro-expression features in the expression video. 
Peng et al. \cite{peng2018} have fine tuned the pre-trained CNN of ImageNet over facial expression datasets for the micro and macro-expression recognition. Li et al. \cite{li2018flow} have applied a 3D flow-based convolutional neural network model for video-based micro-expression recognition. They tried to represent the fine motion flow caused by the minute facial movements using this network.

Efforts have been made to classify micro-expressions using a two-step deep learning architecture \cite{wang2018,kim2016micro,kim2017,khor2018}. In such typical two-step models, the first step extracts spatial features using Convolutional Neural Networks (CNNs) on each frame of the micro-expression video. In the second step, the spatial features  are fed into a Long Short Term Memory (LSTM) based Recurrent Neural Network (RNN) in the same order to learn the temporal correlation between the frames. These methods are unable to learn the spatio-temporal relationship more accurately. Hasani et al. \cite{hasani2017} have introduced another two-step approach to classify the micro-expressions. First, a CNN is used to extract the spatial features from each frame, and then the extracted spatial features are used as the input to linear chain continuous random fields to establish the temporal relation between the frames. Duan et al. \cite{duan2016recognizing} have used the LBP-TOP features along with multiple classifiers from eye-region to recognize the micro-expressions. We have used the eyes and mouth regions in our proposed MicroExpFuseNet model and full face region in our proposed MicroExpSTCNN model for recognizing the micro-expression, and found that, mouth region and some other regions like chicks can depict crucial cues for recognition. Satya et al. \cite{cmu} have used two-stream CNN to learn the spatial and temporal features, respectively. The spatial and temporal features are concatenated to produce the final single feature vector representing the spatio-temporal features which are used for classification by the SVM classifier. Peng et al. \cite{peng2017} have used optical flow sequences obtained from each frame as the input to a 3D CNN architecture for the classification of micro-expressions.
\begin{table}[!t]
\caption{Network Architecture of Proposed MicroExpSTCNN Model. The input dimension is considered for the CAS(ME)$\widehat{}^2$ dataset.}
\centering
\begin{tabular}{c|c|c|c}
\hline
Layer Type&\#Filters&Filter Size&Output Dimension\\
\hline
Input&-&-&64 x 64 x 96\\
3D-Convolution&32 & 3 x 3 x 15 & 32 x 62 x 62 x 82\\
3D-Maxpooling & - & 3 x 3 x 3 & 32 x 20 x 20 x 27\\
Dropout&-&-&32 x 20 x 20 x 27\\
Flatten&-&-&345600\\
Dense&-&-&128\\
Dropout&-&-&128\\
Dense&-&-&3\\
Dropout&-&-&3\\
\hline
\end{tabular}
\label{table:model1}
\end{table}

From the literature review, we can observe that the hand-designed features have the limitations in terms of the robustness and performance in terms of accuracy. Whereas, the deep learning based techniques are the recent trends, showing better accuracy. Still, the use of 3D CNN is limited for micro-expression recognition and uses extra information such as optical flow, etc. Thus, this paper proposes two 3D-CNN methods utilizing the joint spatio-temporal training. The proposed MicroExpSTCNN considers full face as the input, whereas the proposed MicroExpFuseNet considers the eyes and mouth regions as the inputs.

\section{Proposed 3D Spatio-Temporal CNN Models}
Motivated by the success of deep networks in extracting spatio-temporal features for micro-expression recognition, we propose two 3D-CNN models in this paper for micro-expression recognition. The first model named as MicroExpSTCNN is based on the full face regions, whereas the second model named as MicroExpFuseNet is based on the eyes and mouth regions only. Both the proposed models use the 3D-CNN to exploit the joint spatio-temporal relationship. The proposed models are basically the 3D-CNNs while designed with proper care of the number of layers and filter sizes for the facial micro-expression recognition problem. Mainly, this paper focuses over the application of 3D-CNN through proposed MicroExpSTCNN and MicroExpFuseNet for facial micro-expression recognition.

\begin{figure*}[!t]
\includegraphics[width=\textwidth]{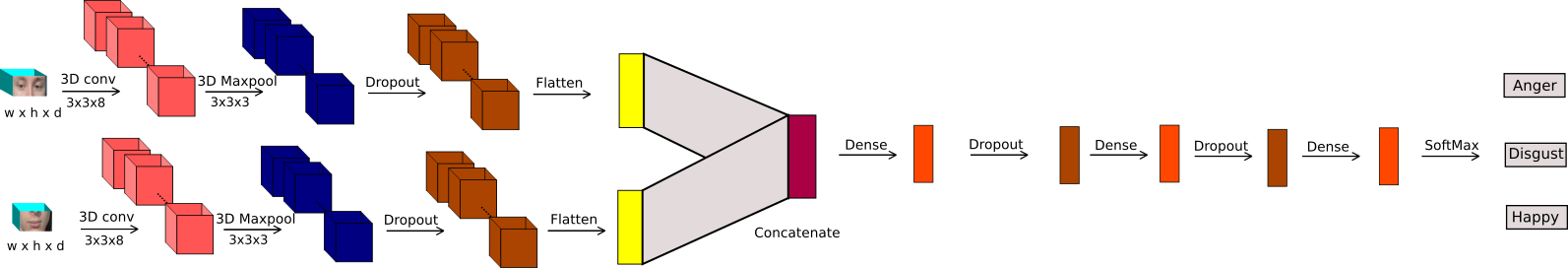}
\caption{The proposed MicroExpFuseNet architecture with the Intermediate fusion scheme for micro-expression recognition using 3D-CNN.}
\label{fig:model2}
\end{figure*}

\begin{figure*}[!t]
\includegraphics[width=\textwidth]{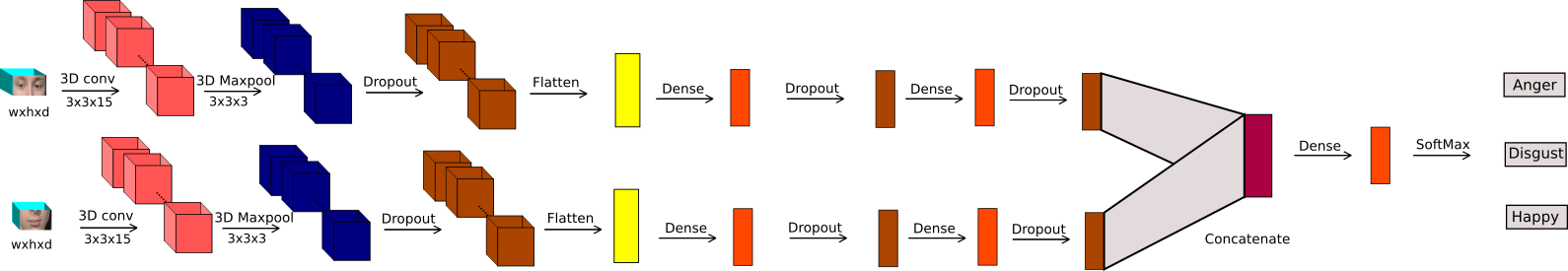}
\caption{The proposed MicroExpFuseNet architecture with the Late fusion scheme for micro-expression recognition using 3D-CNN.}
\label{fig:model3}
\end{figure*}

\subsection{Proposed MicroExpSTCNN Model}
The proposed MicroExpSTCNN is designed to utilize the spatio-temporal features during micro-expression, with utmost priority using 3D-CNN. The MicroExpSTCNN architecture is illustrated in Fig. \ref{fig:model1}. The input dimension to MicroExpSTCNN model is $w \times h \times d$, where $w$ and $h$ are fixed to 64 in this paper and the value of $d$ is dependent upon the dataset used. The proposed MicroExpSTCNN model is composed of 3D convolutional layers, 3D pooling layers, fully connected layers, activation functions and dropouts. The 3D convolutional layers are used to extract the spatial and temporal features by applying the convolution operation using 3D kernel. In contrast to the 2D CNN where the filters are used only in spatial directions, the 3D-CNN uses the filter in temporal direction also.
The 3D pooling layer progressively reduces the dimension output of the 3D convolutional layer while retain the important features. The 3D pooling layer picks the best feature representation in a small spatio-temporal window. The usage of dropout in the network reduces the overfitting of the model over training samples \cite{srivastava2014dropout}. The dropout is used to add the regularization capability in the proposed network. The flatten layer is nothing but the stretching out of multi-dimensional input into a one-dimensional array needed for the fully connected layer.
The dense layers or fully-connected layers are needed to introduce more non-linearity in the network in the form of hierarchical feature extraction. The softmax layer is used to generate the class scores for the classes of the dataset being used.

Our proposed network is composed of stacking, one 3D convolutional layer with 32 filters of dimension 3 x 3 x 15, one 3D pooling layer with a kernel size 3 x 3 x 3 and two fully-connected (dense) layers. The final dimension of fully-connected layer depends on the number of expression labels in the dataset. Table \ref{table:model1} summarizes the network architecture of the proposed MicroExpSTCNN model in terms of the filter dimension and the output dimension of different layers. The input dimension in Table \ref{table:model1} is considered for the CAS(ME)$\widehat{}^2$ dataset. The output dimension of final dense layer varies for different datasets.

\begin{table}[t!]
\caption{Network Architecture of Proposed Intermediate MicroExpFuseNet Model. The input dimension is considered for the CAS(ME)$\widehat{}^2$ dataset. The number of nodes in the final dense layer depends upon the number of classes in a dataset.}
\centering
\begin{tabular}{c|c|c}
\hline
Layer Type & Filter Size & Output Dimension\\
\hline
Input - 1 & - & 32 x 32 x 96\\
Input - 2 & - & 32 x 32 x 96\\
3D-Convolution - 1 & 3 x 3 x 15 & 32 x 30 x 30 x 82\\
3D-Convolution -2 & 3 x 3 x 15 & 32 x 30 x 30 x 82\\
3D - Maxpooling - 1 & 3 x 3 x 3 & 32 x 10 x 10 x 27\\
3D - Maxpooling - 2 & 3 x 3 x 3 & 32 x 10 x 10 x 27\\
Dropout - 1 & - & 32 x 20 x 20 x 27\\
Dropout - 2 & - & 32 x 20 x 20 x 27\\
Flatten - 1 & - & 86400\\
Flatten - 2 & - & 86400\\
Concatenate & - & 86400\\
Dense & - & 1024\\
Dropout & - & 1024\\
Dense & - & 128\\
Dropout & - & 128\\
Dense & - & 3\\
\hline
\end{tabular}
\label{table:model2}
\end{table}

\subsection{Proposed MicroExpFuseNet Model}
The proposed MicroExpSTCNN described in the previous sub-section considers the whole face region as the input. However, it is observed by the researchers that the eyes and mouth regions contribute more towards the expression analysis, compared to the other regions of the face \cite{duan2016recognizing,iwasaki2016hiding,agrawal2014emotion}. Eyes region are used for feature extraction in Duan et al. \cite{duan2016recognizing}. Iwasaki et al. have analyzed the simultaneous eyes and mouth movement correlation \cite{iwasaki2016hiding}. Agrawal et al. have exploited the left eye, right eye and mouth regions for extracting different hand-designed features and fed into an SVM \cite{agrawal2014emotion}. Considering only eyes and mouth region leads to a computationally efficient model. Thus, in this paper, we propose a region-based 3D-CNN model (i.e., MicroExpFuseNet model). In MicroExpFuseNet model, only eyes and mouth regions of the face are used as input to two separate 3D spatio-temporal CNNs. Both CNNs are later fused and converged into a single network. We perform the preprocessing over each frame in the expression video to detect the eyes and mouth regions using DLib face detector\footnote{\url{dlib.net/face_detector.py.html}}, which is applied to detect and align face in each frame by first detecting the 68 landmarks in the face. These landmarks are used to crop the eyes and mouth regions. Based on the different fusion strategies (i.e., at different stages), we propose two versions of MicroExpFuseNet models: Intermediate MicroExpFuseNet and Late MicroExpFuseNet.

\subsubsection{Intermediate  MicroExpFuseNet  Model}
In Intermediate MicroExpFuseNet Model, as the name suggests, the features of two 3D convolutional neural networks (3D-CNN) are fused at some intermediate level. The eye portion (including both eyes) of the face is fed as the input to one of the 3D-CNN and mouth portion is fed as the input to another 3D-CNN. The features extracted from both eye and mouth regions are fused together at some intermediate level. The proposed Intermediate MicroExpFuseNet architecture is illustrated in Fig. \ref{fig:model2}.

The proposed MicroExpFuseNet model has two separate 3D-CNN with each network composed by stacking, one 3D convolutional layer with 32 filters of size 3 x 3 x 15, one 3D pooling layer with a kernel size of 3 x 3 x 3 similar to the MicroExpSTCNN model. The flatten layer is used to convert the activation map into the single dimension feature vector. The flattened features from both networks are concatenated to form a new vector. In Intermediate MicroExpFuseNet, the fused features are again processed with dense and dropout layers before class score generation. Table \ref{table:model2} reports the network architecture of the proposed Intermediate MicroExpFuseNet model in terms of the filter dimension and the output dimension of different layers. In Table \ref{table:model2}, the input dimension is considered for the CAS(ME)$\widehat{}^2$ dataset.

\begin{table}[!t]
\caption{Network Architecture of Proposed Late MicroExpFuseNet Model. The input dimension is considered for the CAS(ME)$\widehat{}^2$ dataset. The number of nodes in the final dense layer depends upon the number of classes in a dataset.}
\centering
\begin{tabular}{c|c|c}
\hline
Layer Type & Filter Size & Output Dimension\\
\hline 
Input - 1 & - & 32 x 32 x 96\\
Input - 2 & - & 32 x 32 x 96\\
3D-Convolution - 1 & 3 x 3 x 15 & 32 x 30 x 30 x 82\\
3D-Convolution -2 & 3 x 3 x 15 & 32 x 30 x 30 x 82\\
3D - Maxpooling - 1 & 3 x 3 x 3 & 32 x 10 x 10 x 27\\
3D - Maxpooling - 2 & 3 x 3 x 3 & 32 x 10 x 10 x 27\\
Dropout - 1 & - & 32 x 10 x 10 x 27\\
Dropout - 2 & - & 32 x 10 x 10 x 27\\
Flatten - 1 & - & 86400\\
Flatten - 2 & - & 86400\\
Dense - 1 & - & 1024\\
Dropout - 1& - & 1024\\
Dense - 2 & - & 1024\\
Dropout - 2& - & 1024\\
Dense - 3 & - & 128\\
Dropout - 3& - & 128\\
Dense - 4 & - & 128\\
Dropout - 3& - & 128\\
Concatenate& - & 256\\
Dense & - & 3\\
\hline
\end{tabular}
\label{table:model3}
\end{table}

\subsubsection{Late MicroExpFuseNet Model}
In Late MicroExpFuseNet Model, as the name suggests, the features of two 3D-CNN are fused just before the final dense layer. In this model the eye portion of the face is fed as the input to one of the 3D CNN and the mouth portion is fed as the input to another 3D CNN. The features extracted from both eye and mouth regions are fused together at the last fully-connected layer. The proposed Late MicroExpFuseNet architecture is shown in Fig. \ref{fig:model3}.
This model also has two separate 3D CNNs with each CNN composed of stacking, one 3D convolutional layer with 32 filters of size 3 x 3 x 15, one 3D pooling layer with a kernel size of 3 x 3 x 3 and a flatten layer to achieve a single dimension feature vector. The dropout, flatten and dense layers are used in both networks. Both networks are fused before final dense layer. Table \ref{table:model3} presents the network architecture of the proposed Late MicroExpFuseNet model in terms of the filter dimension and the output dimension of different layers. In Table \ref{table:model3}, the input dimension is considered for the CAS(ME)$\widehat{}^2$ dataset.

Note that, for both Intermediate and Late MicroExpFuseNet models, the output dimension of the final dense layer is different for different datasets.

\begin{table}[!t]
\caption{The summary of the micro-expression video datasets (i.e., CAS(ME)$\widehat{}^2$ and SMIC datasets) in terms of the type of expressions present and number of video samples. In this table, the number of samples in a class for a dataset is mentioned as X=Y+Z, where X, Y and Z are the total number of samples, the number of samples in the training set and the number of samples in validation set, respectively.}
\centering
\begin{tabular}{c|c|c}
\hline
Expression & CAS(ME)$\widehat{}^2$ & SMIC\\
\hline
Happy & 94=73+21 & -\\
Angry & 76=64+12 & -\\
Disgust & 36=28+8 & -\\
Negative & - & 79=60+19\\
Positive & - & 34=28+6\\
Surprise & - & 43=36+7\\

\hline
Total & 206=165+41 & 156=124+32 \\
\hline
\end{tabular}
\label{table:dataset}
\end{table}

\begin{table*}[!t]
\centering
\caption{Results comparison in terms of the accuracy of proposed MicroExpSTCNN and MicroExpFuseNet with state-of-the-results methods. The reported accuracies of various hand-crafted methods (HCM) and deep learning methods (DLM) for micro expression recognition over CAS(ME)$\widehat{}^2$ and SMIC datasets. The results of compared methods are taken from its source papers. Note that the training and test set vary for the different methods. The best result for a dataset is highlighted in bold.}
\begin{tabular}{c|c|c|c|c c c}
\hline
Method & Proposed Year & Method Type & CAS(ME)$\widehat{}^2$ & SMIC\\
\hline
LBP-TOP \cite{li2013spontaneous} & 2013 & HCM & - & 42.72\%\\
STCLQP \cite{huang2016spontaneous} & 2016 & HCM & - & 64.02\%\\
\hline
CNN with Augmentation \cite{takalkar2017} & 2017 & DLM & 78.02\% & -\\
3D-FCNN \cite{li2018flow} & 2018 & DLM & - & 55.49\%\\
\hline
MicroExpSTCNN & Proposed & DLM & \textbf{87.80\%} & \textbf{68.75\%}\\
MicroExpFuseNet (Intermediate) & Proposed & DLM & 83.25\% & 54.77\%\\
MicroExpFuseNet (Late) & Proposed & DLM & 79.31\% & 64.82\%\\
\hline
\end{tabular}
\label{table:comparison}
\end{table*}

\begin{table}[!t]
\caption{The confusion matrix using MicroExpSTCNN model over CAS(ME)$\widehat{}^2$ dataset.}
\centering
\begin{tabular}{c|c|c|c}
\cline{1-4}
Class & Happy & Angry & Disgust \\
\cline{1-4}
Happy & 19 & 1 & 1 \\
Angry & 0 & 11 & 1 \\
Disgust & 1 & 1 & 6 \\
\cline{1-4}
\end{tabular}
\label{table:confusion_matrix1}
\end{table}

\begin{table}[!t]
\caption{The confusion matrix using MicroExpSTCNN model over SMIC dataset.}
\centering
\begin{tabular}{c|c|c|c}
\cline{1-4}
Class & Negative & Positive & Surprise \\
\cline{1-4}
Negative & 14 & 0 & 5 \\
Positive & 2 & 2 & 2 \\
Surprise & 1 & 0 & 6 \\
\cline{1-4}
\end{tabular}
\label{table:confusion_matrix2}
\end{table}

\section{Experimental Setting}
This section discusses the experimental settings followed in this work. First, we provide a brief overview of each micro-exoression dataset used for the experiments, followed by a thorough discussion over the hyper-parameter settings used for the training of the proposed networks in different subsections.

\subsection{Micro-Expression Datasets Used}
This sub-section provides brief descriptions of the datasets used in this study for micro-expression recognition. One of the major problems with any deep learning based techniques is the requirement of a sufficient size of data. In order to meet the requirement, we have used the benchmark micro-expression video datasets such as CAS(ME)$\widehat{}^2$ \cite{qu2017cas} and SMIC \cite{li2013spontaneous} in our study. Table \ref{table:dataset} shows the expression levels and the number of video samples present in the respective datasets. We have used 80\% of data from each dataset for training and other 20\% for validation. Note that the training and validation splitup is done once and then same training and validation sets are used for all the experiments. We have followed the whole dataset splitup with varying \% of samples in different classes. The followings are the details of the micro-expression datasets used in this paper.

\begin{enumerate}
\item CAS(ME)$\widehat{}^2$ Database \cite{qu2017cas}:
CAS(ME)$\widehat{}^2$ is the latest version of the CASME series of datasets on facial micro-expressions containing 206 videos. This dataset contains 3 classes. Note that we have only used micro-expression videos that have more than 96 frames to maintain the consistency over the data.
\item Spontaneous Micro Expression Database (SMIC) \cite{li2013spontaneous}:
SMIC is a dataset of spontaneous micro-expressions rather than posed micro expressions. In this paper, we have used only High Speed (HS) profile of SMIC dataset which consists of a total of 156 micro-expression videos of three categories: surprise, positive and negative. The other profiles of the SMIC dataset such as Visual Normal (VIS) and Near Infra-Red (NIR) are not used as they are not suited with deeper 3D filters. Note that we have used only those micro-expression videos that have more than 18 frames to maintain the consistency over the data.
\end{enumerate}

\subsection{Hyper-Parameters Settings}
We have implemented our model on Keras with Tensorflow at the backend. The model was trained and tested on a GPU Server with NVIDIA Tesla K80 24GB GDDR5 graphics processor. Both MicroExpSTCNN and MicroExpFuseNet, use Categorical Cross Entropy loss function and SGD optimization technique with the default learning rate schedule. Moreover, both the networks are trained for 100 epochs with batch size of 8. The input dimensions for MicroExpSTCNN and MicroExpFuseNet models are 64 x 64 x 96 and 32 x 32 x 96, respectively.

Next we discuss the results obtained by the proposed architectures compared to the state-of-the-art.

\section{Experimental Results and Discussions}
The experimental results along with a rigorous comparison with the state-of-the-art methods are presented in this section. The accuracy standard deviation analysis is also conducted for the proposed methods to analyze the stability of the model. The impact of different 3D kernel sizes and the impact of facial features are also analyzed. 

\subsection{Experimental Results}
The results comparison in terms of the accuracy of proposed MicroExpSTCNN and MicroExpFuseNet (Intermediate and Late) methods with the state-of-the-art hand-crafted methods (HCM) and deep learning methods (DLM) is shown in Table \ref{table:comparison} over CAS(ME)$\widehat{}^2$ and SMIC datasets. Note that the results of compared methods are taken from the corresponding source papers. It can be observed that the proposed MicroExpSTCNN method outperforms other methods over CAS(ME)$\widehat{}^2$ and SMIC datasets. 
The confusion matrix using the proposed MicroExpSTCNN model over CAS(ME)$\widehat{}^2$ and SMIC datasets are also presented in Table \ref{table:confusion_matrix1} and Table \ref{table:confusion_matrix2}, respectively. 
The performance of proposed MicroExpFuseNet models (especially Intermediate one) is better than most of the existing methods. The utilization of spatio-temporal joint training using 3D CNN in both proposed MicroExpSTCNN and MicroExpFuseNet models facilitate the more accurate feature extraction for the micro-expression recognition. Among the proposed methods, the performance of MicroExpSTCNN model is better than MicroExpFuseNet models. The possible reason for this observation is due to the facial features which is analyzed in the below subsection ``Impact of Facial Features''. It is also observed that the Intermediate fusion is better suitable as compared to the Late fusion in the MicroExpFuseNet model over CAS(ME)$\widehat{}^2$ dataset. Whereas, the Late fusion is preferable over SMIC dataset with MicroExpFuseNet model.

\begin{table}[!t]
\centering
\caption{The mean accuracy with standard deviation for the proposed MicroExpSTCNN and MicroExpSTCNN models over CAS(ME)$\widehat{}^2$ and SMIC datasets. The Standard Deviation is computed from 91 to 100 epochs to show the stability of models in last few epochs.}
\centering
\begin{tabular}{c|c|c|c}
\hline
\multirow{2}{*}{Database} & \multirow{2}{*}{MicroExpSTCNN} & \multicolumn{2}{c}{MicroExpFuseNet} \\
\cline{3-4}
& & Intermediate Fusion & Late Fusion\\
\hline
CAS(ME)$\widehat{}^2$ & 82.20$\pm$5.02\% & 78.57$\pm$5.78\% & 73.20$\pm$3.87\% \\
SMIC & 62.50$\pm$2.55\% & 51.20$\pm$3.56\% & 59.95$\pm$3.19\% \\
\hline
\end{tabular}
\label{table:ourresults}
\end{table}

\subsection{Accuracy Standard Deviation Analysis}
We have reported the standard deviation in the results along with the mean accuracy to highlight the stability of the model, even for the less training data. Table \ref{table:ourresults} reports the mean and standard deviation in validation accuracies between epochs 91 to 100 for MicroExpSTCNN and MicroExpFuseNet models over CAS(ME)$\widehat{}^2$ and SMIC datasets. 
It is observed that the proposed MicroExpSTCNN and MicroExpFuseNet models exhibit the high stability over SMIC datasets, whereas it is reasonable over CASME$\widehat{}$~2.

\begin{table}[!t]
\caption{The performance of the MicroExpSTCNN model in terms of accuracy with varying filter sizes over CAS(ME)$\widehat{}^2$ dataset. The best result is highlighted in Bold.}
\centering
\begin{tabular}{c|c|c|c}
\hline
Filter Size & Accuracy [\%] & Filter Size & Accuracy [\%]\\
\hline
3 x 3 x 3 & 82.93 & 5 x 5 x 15 & 51.22\\
3 x 3 x 7 & 80.49 & 5 x 5 x 19 & 29.27\\
3 x 3 x 15 & \textbf{87.80} & 7 x 7 x 3 & 70.73\\
3 x 3 x 19 & 29.27 & 7 x 7 x 7 & 51.22\\
5 x 5 x 3 & 63.30 & 7 x 7 x 15 & 29.27\\
5 x 5 x 7 & 58.54 & 7 x 7 x 19 & 51.22 \\
\hline
\end{tabular}
\label{table:filter}
\end{table}

\subsection{Impact of 3D Kernel}
The original architecture of the MicroExpSTCNN model is constructed around the idea of 3D-convolution. The performance of the model is dependent on many hyper-parameters of the network. One of the important hyper-parameter is the dimension of the filter used for the 3D-convolution. It is desired to analyze the impact of 3D kernel size as it depicts the extent of spatial and temporal exploitation in the process of feature extraction. So, we have conducted the experiments to find the optimal filter size that achieves the best performance. Table \ref{table:filter} illustrates the model accuracy on varying 3D filter sizes over CAS(ME)$\widehat{}^2$ dataset using MicroExpSTCNN model. The optimal 3D kernel size is found to be 3 x 3 x 15. It is observed that the filter with less spatial extent and more temporal extent is better suitable. It is also seen that the deeper kernel works better with smaller filters.

\subsection{Impact of Facial Features}
\begin{figure}[!t]
\centering
\includegraphics[width=\columnwidth]{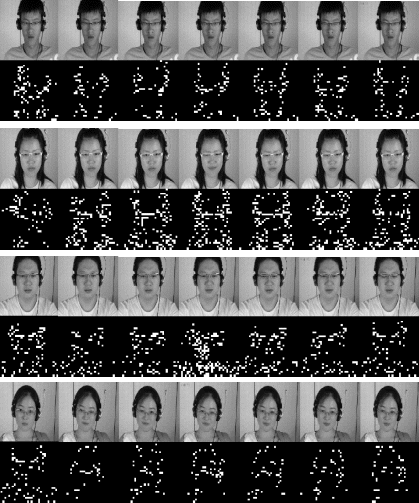}
\caption{The impact of different facial features with the help of saliency maps. The saliency map is binarized for better visualization. It can be observed that the facial features apart from Eyes and Mouth regions also contribute towards the micro-expression recognition.}
\label{model:salience}
\end{figure}

In this section, we analyze results of some experiments to understand the reason for better performance of the MicroExpSTCNN model compared to the Intermediate and Late MicroExpFuseNet models. It is evident that the performance of a model is dependent on the features that it learns while training. We have built our MicroExpFuseNet models by assuming that the eyes and mouth regions contribute heavily to facial micro-expression recognition. Whereas, MicroExpSTCNN model performs better than MicroExpFuseNet Models. Thus, in order to find out the important facial features for micro-expression recognition, we have analyzed the saliency maps. The saliency maps are computed to gain deeper understanding about the relevant facial features in a micro expression by measuring the positive response towards the class scores. The saliency maps along with the original 7 frames at different times for 4 different videos are illustrated in Fig. \ref{model:salience}. The $2^{nd}$, $4^{th}$, and $8^{th}$ rows are the saliency maps for the frames shown in $1^{st}$, $3^{rd}$, $5^{th}$, and $7^{th}$ rows, respectively. It is observed from these salience maps that other facial features apart from the eyes and mouth are also important to classify the facial micro-expressions. Thus, this is one of the possible reasons that the performance of the MicroExpSTCNN model is better than the MicroExpFuseNet models. 

\section{Conclusion}
This paper proposes two 3D Convolutional Neural Networks (CNNs) for micro-expression recognition from videos. The 3D-CNN is used for the joint convolution in spatial and temporal directions which leads to the simultaneous spatio-temporal training. Our rigorous experiments on the two proposed CNNs over the benchmark datasets show that, other than eyes and mouth regions, some other salient facial regions can contribute to micro-expression recognition. 
We have also noticed the stability of models over SMIC dataset, whereas it is reasonable over CAS(ME)$\widehat{}^2$ dataset. The effect of 3D kernel size is also analyzed and concluded that the smaller spatial and deeper temporal filters are better suitable. In the future, we can extend this work by experimenting with different salient points on the face and analyzing the key facial regions to recognize micro-expressions.

{\small
\bibliographystyle{IEEEtran}
\bibliography{Reference}}

\end{document}